\documentclass{article}

\usepackage{arxiv}
\usepackage{cite}
\usepackage{multirow}
\usepackage{amsmath,amssymb,amsfonts}
\usepackage{algorithmic}
\usepackage{graphicx}
\usepackage{textcomp}
\usepackage{xcolor}
\usepackage{url}

\def\BibTeX{{\rm B\kern-.05em{\sc i\kern-.025em b}\kern-.08em
    T\kern-.1667em\lower.7ex\hbox{E}\kern-.125emX}}

\begin{document}

\title{Can LLMs Assist  Expert Elicitation for Probabilistic Causal Modeling?}


\author{
 Olha Shaposhnyk, Daria Zahorska, and Svetlana Yanushkevich \\
  Biometric Technologies Laboratory, Schulich School of Engineering\\
  University of Calgary, Canada\\
  \texttt{olha.shaposhnyk1@ucalgary.ca} \\
  }

\maketitle

\begin{abstract}
Objective: 
This study investigates the potential of Large Language Models (LLMs) as an alternative to human expert elicitation for extracting structured causal knowledge and facilitating causal modeling in biometric and healthcare applications.

Material and Methods:
LLM-generated causal structures, specifically Bayesian networks (BNs), were benchmarked against traditional statistical methods (e.g., Bayesian Information Criterion) using healthcare datasets. Validation techniques included structural equation modeling (SEM) to verifying relationships, and measures such as entropy, predictive accuracy, and robustness to compare network structures. 

Results and Discussion:
LLM-generated BNs demonstrated lower entropy than expert-elicited and statistically generated BNs, suggesting higher confidence and precision in predictions. However, limitations such as contextual constraints, hallucinated dependencies, and potential biases inherited from training data require further investigation. 

Conclusion:
LLMs represent a novel frontier in expert elicitation for probabilistic causal modeling, promising to improve transparency and reduce uncertainty in the decision-making using such models. 
\end{abstract}


\section{Introduction}

Probabilistic causal graph models \cite{Pearl} are the core of the decision support systems in smart health, which rely on multi-domain data (e.g. time-series data, statistics, demographics). These models, such as Structural Equation Models (SEM) and Bayesian networks (BNs), capture causality between data patterns and variables by representing probabilistic dependencies among them. 

Constructing probabilistic causal graphs typically relies on human experts \cite{Pearl}. This process is time-consuming, subjective, and often constrained by domain-specific expertise availability. 
In contrast, there are multiple data-driven approaches, also called statistical causal discovery. Those models are often inaccurate due to the lack of domain knowledge brought by experts in the aforementioned expert-built causal graphs \cite{Glymour}. 
  
Most recently, rapid progress in the development of Large Language Models
(LLMs) prompted the studies on the usage of LLMs for building prior distributions for causal graph discovery \cite{Darvariu2024} as well as for statistical causal discovery \cite{Takayama}.

 On the other hand, in the domain of AI-assisted medical diagnostics, there is significant progress in calibrating subjective or imperfect expert knowledge through expert elicitation \cite{[Bojke-2021]}.

Expert elicitation is the process of soliciting probabilistic beliefs from human (or AI) experts about unknown quantities or parameters.  Multiple experts are needed to capture the variety of backgrounds, opinions and knowledge provided in the form of distributions from the experts. Elicitation of expert knowledge involves defining the target questions, creating an elicitation protocol, training the experts in subjective probability, conducting the interviews, providing feedback to the experts, and analyzing the results \cite{[Cooke-1991],[Mayer_Elicitation_RAND_2021]}. 

 In this study, we utilize LLMs as a tool for expert elicitation to extract causal relationships and construct Bayesian networks (BNs). This approach is applied to a case study on 'Sleep, Health and Lifestyle' within the smart health domain. The LLM-generated BNs are compared against those produced using established methods.

{This paper is structured as follows: Section II presents the most important related works.  Section III provides our motivation and contribution. Section IV outlines the data and methodology applied to this study. The experimental study and the results are described in Sections V and VI.  Section VII concludes the paper.}

\section{Related Works}




Probabilistic causal models such as BNs \cite{Pearl} are prevalent in multiple diagnostics and inferential analysis due to their ability to model uncertainty and facilitate interpretable probabilistic reasoning, making them valuable for decision support and risk assessment. 

BNs are performing with a high level of accuracy in diagnosing and forecasting disease \cite{Polotskaya}. Nearly two-thirds of BN-based studies have concentrated on four major categories of clinical diagnostics: cardiac, cancer, psychological, and lung disorders, as indicated in \cite{McLachlan}.

In \cite{Shaposhnyk}, BNs were applied to infer cognitive load from a combination of physiological signals and personality traits. Their approach illustrates the ability of BNs to represent latent cognitive states through the integration of heterogeneous features, providing personalized insights that are crucial for applications in clinical decision support and occupational health. BNs were used in \cite{Fenton}  for the assessment of cognitive biases of expert elicitation, such as ambiguity effect, attentional bias, confirmation bias,  framing effect,  and expectation bias.


The BN is a probabilistic causal structure, which is Directed Acyclic Graph (DAG). It is primarily constructed using human expert knowledge. There are several automated data-driven statistical methods for BN structure learning: score-based methods such as Bayesian Information Criterion (BIC) {\cite{Schwarz}} and Akaike Information Criterion (AIC) {\cite{Akaike}}, constrained-based tests such as Peter-Clark (PC) Algorithm {\cite{Spirtes}} and Fast Causal Inference (FCI) Algorithm) {\cite{Spirtes-2001}}. Score-based methods evaluate candidate network structures using a scoring function that balances model fit and complexity. The mentioned criteria penalize model complexity to prevent overfitting. However, these algorithms are computationally expensive, because they have to enumerate and score every possible graph among the given variables \cite{Nogueira}. Constrained-based methods do not rely on a predefined structure and build the network by analyzing conditional independence between variables in the dataset \cite{Versteeg}; they are, however, inaccurate and prone to errors.

  A combination of the classical constrained-based methods (such as the PC algorithm and the Linear Non-Gaussian Acyclic Model (LiNGAM)) and LLMs was proposed in  \cite{Khatibi} to create causal structures. They proposed a multi-component Autonomous LLM-Augmented Causal Discovery Framework (ALCM), which allows one to build a causal graph using classical methods and then apply a causal wrapper to encapsulate and translate the raw initial causal graph into a series of contextual, causal-aware prompts. The prompts are refined by LLM to form a coherent graph. The accuracy {of ALCM in learning causal graph structures ranges from} 65\% to 98\%, showing great performance on both small and large benchmark datasets. 

In \cite{Takayama}, causal graphs were first built using PC, Exact Search and DirectLiNGAM algorithms and then refined by applying LLMs. The statistical causal prompting process integrated two prompts: one for generating the domain knowledge with a given causal structure and bootstrap probabilities, and the other to objectively judge the existence of a causal effect between variables, responding "yes" or "no" to indicate the causality using GPT-4. 

Despite the promise of LLM-enhanced causal discovery, the key limitation in these studies is the lack of demonstration of its efficiency for real-world deployment scenarios. Most frameworks are evaluated on benchmark datasets, and their practical applicability in the areas where domain expertise, noisy data, and ethical constraints are central is not yet fully addressed.

\section {Problem Formulation and Contribution}

  This study addresses the following question: Can LLMs serve as a viable tool for expert elicitation in constructing causal models such as Bayesian networks (BNs)?
	
 We apply this approach to a decision-support case study analyzing  'Sleep, Health and Lifestyle' data in the smart health domain. We demonstrate the utility of the LLM-generated BN for diagnostics, inference, and scenario analysis in healthcare decision-making. Performance is evaluated by comparing LLM-derived BNs against those generated using established statistical methods (e.g., BIC, PC algorithms).
 

 The existing approaches to building BNs include deploying human expert knowledge, data-driven approaches, and statistical methods (represented as Stats. node) based on BIC, PC, or SEM, as shown in the diagram below. In this work, we consider conducting expert knowledge elicitation using LLMs foe causal structure discovery (highlighted in green):

\begin{equation*}\label{fig:ensemble-machine-technique}
		\texttt{\small Data}\hspace{-3mm}
	\mbox{
	\begin{tabular}{clc} 
	$\nearrow$   &\hspace{-3mm}\texttt{\small Expert}   &\hspace{-3mm}$\rightarrow$\\
    $\rightarrow$   &\hspace{-3mm}\texttt{\small Stats.}   &\hspace{-3mm}$\rightarrow$\\
	$\searrow$   &\hspace{-3mm}\texttt{\small \textcolor{green}{LLM}}    &\hspace{-3mm}$\rightarrow$
	\end{tabular}
	\hspace{-3mm}
\begin{tabular}{clcc} 
\hspace{-3mm}\texttt{\small BN I}    
&\hspace{-3mm}$\searrow $   &\\
\hspace{-3mm}\texttt{\small BN II} 
&\hspace{-3mm}$\rightarrow$ &\\
\hspace{-3mm}\texttt{\small \textcolor{green}{BN III}}   
&\hspace{-3mm}$\nearrow$    &\\
	\end{tabular}
	} 
	\hspace{-7mm}\texttt{\small Decision Support}
\end{equation*}

This approach enables a better understanding of causal relationships in the resulting structure, refined through LLM-aided analysis. In other words, the LLM functions as a proxy for one or more experts, building upon their diverse knowledge. We evaluate its efficacy in a case study to determine whether LLMs can support or optimize expert elicitation for constructing causal models.



\section{Methodology}

This section describes the methodology used to construct and evaluate causal models using LLM for expert elicitation.

\begin{figure*}
    \centering
    \includegraphics[width=0.9\linewidth]{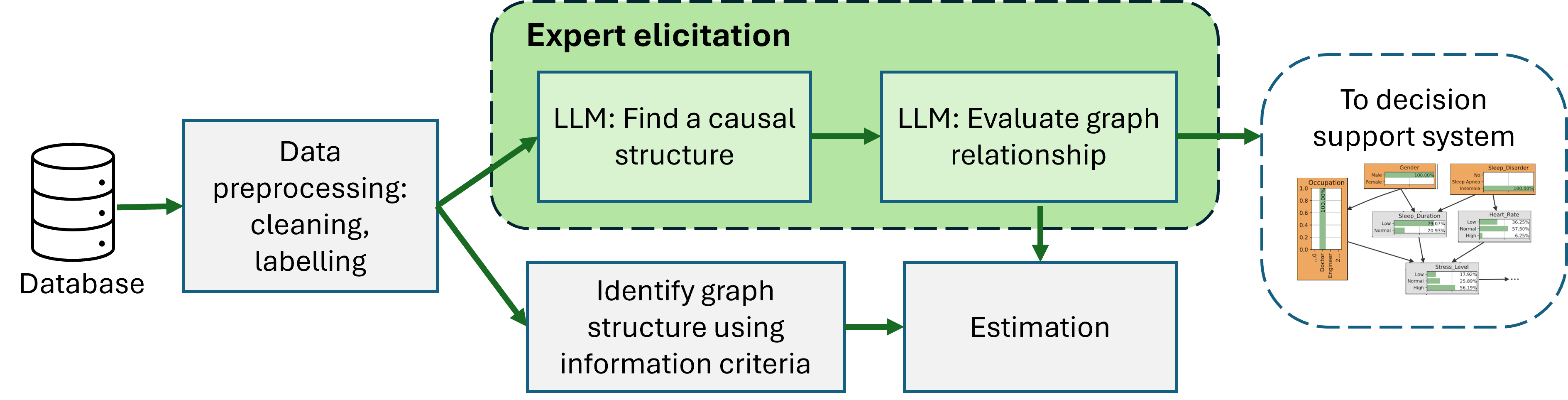}
    \caption{Causal modeling workflow: data preprocessing, building causal graph using LLM and information criteria, and constructing a BN. The novelty of this work such as applying the LLM-based expert elicitation is highlighted in green.}
    \label{fig:workflow}
\end{figure*}

Our approach is illustrated in Figure \ref{fig:workflow}. The experimental design involves data preparation, creation of the causal graph models, including the proposed approach expert elicitation through LLMs, and verification of the graph structure. The created structure is used in the decision support with the BN as a core. 

\subsection{Data selection}

In this study, we used the \textit{Sleep Health and Lifestyle Dataset} \cite{Dataset}, which comprises 400 rows and 13 columns and includes variables such as sleep duration, stress level, physical activity, etc. To increase the data quality, we applied a preprocessing pipeline that included data cleaning and value labeling.
The following steps were taken to clean the data:
\begin{itemize}
    \item Removing duplicates to prevent bias in causal relationship discovery,
    \item Outlier detection and removal, and
    \item Value labelling to split continuous data into groups.
\end{itemize}

To address the discretization of continuous data,  the numerical variables were categorized into groups based on domain knowledge as follows:
\begin{itemize}
    \item Quality of Sleep  $\rightarrow$  Bad, Normal, Good.
    \item Stress level  $\rightarrow$ Low, Moderate, High.
\end{itemize}
\subsection{BN Construction}

BN consists of a DAG where nodes represent random variables and edges represent the relationship between them, and a conditional probability distributions associated with each of the random variables. BNs reflect both statistical distribution learned from the data and domain expertise. It allows to perform inference to examine various scenarios, querying the marginals of states given some observations made anywhere in the network; the marginal distributions of nodes are updated and their impact is propagated through to the node of interest.

To construct the BN, researchers utilize a related causal structure, SEM, which evaluates relationships between observed and latent variables. Following the methodology in \cite {Shaposhnyk-2023}, we  employ SEM to validate variable relationships before their inclusion in the BN.

\subsection{Expert elicitation for constructing BN}

{In both the human expert-based and data-driven BN construction,  a DAG structure is created using the variables available in the data set. Validation of the 'quality' of cthe reated structure is performed using the information criteria described below.}

These criteria include a Multivariate information-based Inductive Causation (MIIC) \cite{Verny2017} and the BIC to identify potential causal relationships between variables. These methods generate possible causal relationships, represented as directed edges in the DAG. However, while these methods provide statistically significant results, they may lack domain-specific context or fail to account for unobserved factors. 

To address this limitation, we used an LLM as a framework for expert elicitation to discover causal relationships and find possible confounders in the graph structure. The LLM prompts were created in order to assess the relationships found based on domain knowledge and provide reasoned explanations for their validity or invalidity. We deployed a state-of-the-art prompt engineering strategy \cite{Zamfirescu} to improve output quality.

 By applying multiple LLMs in sequence, we   {verified consistency across models}, discovered relationships and identified possible misinterpretations. For example, if the first LLM suggested a causal link between variables, the second LLM examined and verified whether this relationship aligned with deeper domain knowledge and highlighted potential confounding variables.

To further validate these relationships within our dataset \cite{Shaposhnyk}, we employed an SEM building tool \cite{Igolkina}. In the final step, the refined causal relationships were used to construct a BN.  This approach ensured that the resulting BN effectively represents both statistical patterns in the data and domain-informed causality. Note that the created BN is intended to be a core tool for the decision support as described in section \ref{sec:DS}.

\section{Causal Modeling }

In this section, we apply different methods to create a causal graph structure and compare the results obtained by traditional data-driven statistical techniques, human expert knowledge-based, and using LLM as expert elicitation.

\subsection{Causal graph structure created by human experts}

\begin{figure}
    \centering
    \includegraphics[width=1\linewidth]{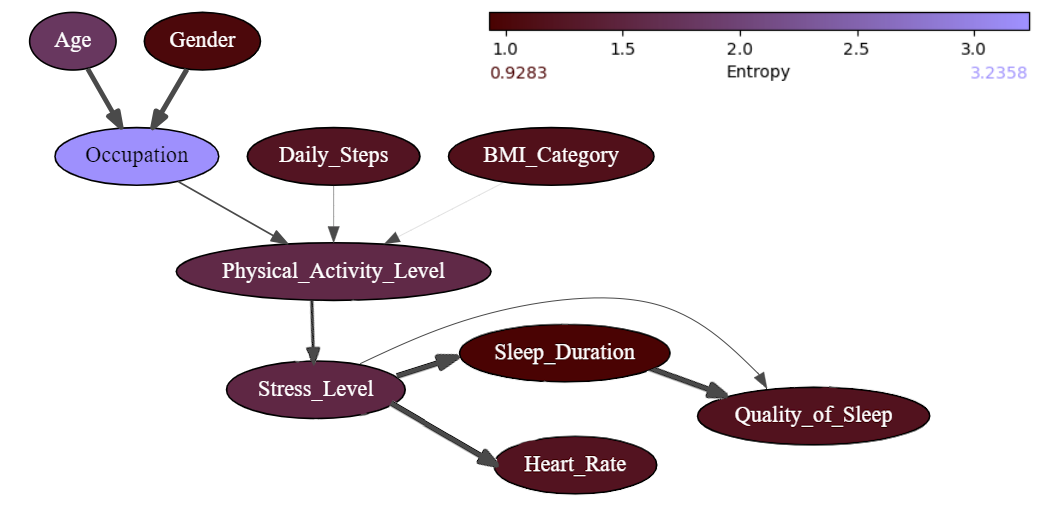}
    \caption{Causal graph structure derived from human expertise. Node color indicates entropy values computed per node, while arc thickness represents mutual information calculated for each connection. }
    \label{fig:expert_graph}
\end{figure}

An example of a BN created by human experts with limited knowledge for the case study of the dataset \textit{Sleep Health and Lifestyle Dataset} \cite{Dataset} is shown in Figure \ref{fig:expert_graph}.

{The BN consists of 10 nodes representing the data variables. To validate the relationships between them, we employed SEM. This method allows us to assess the strength and direction of causal relationships by estimating regression coefficients (estimate) and testing their significance.
}

{Our SEM analysis provides that all relationships were statistically significant, except for two, (\textit{Physical activity} $\rightarrow$ \textit{Stress level}, and \textit{Stress level} $\rightarrow$ \textit{Heart rate}). These relationships did not reach statistical significance, with $p$-values exceeding 0.05. This suggests that either these relationships may be reversed, or they may not exist at all in the context of our data.}

\subsection{Causal graph structure based on information criteria}

The existing techniques for generating a causal graph directly from the data are based on statistical methods  { \cite{Glymour} }and information criteria  { \cite{bollen2014, de2018}}. This approach leverages algorithms such as the MIIC and the BIC to estimate causal relationships based on patterns in the data \cite{Cabeli}. MIIC provides an initial causal structure by identifying statistical dependencies and orienting edges based on mutual information and independence tests.
BIC refines and validates the structure by scoring candidate graphs and selecting the one that optimally balances likelihood and complexity.

MIIC is particularly effective in reconstructing causal structures from observational data, accounting for latent confounders and indirect dependencies, including the following steps: 
\begin{enumerate}
    \item Computation of mutual information to quantify the statistical dependence between pairs of variables, measuring the amount of shared information.
    \item  MIIC calculation for refining the relationships   {between pairs of variables}   to distinguish direct and indirect dependencies through conditional mutual information,.
    \item Identifies potential latent variables that influence observed relationships.
\end{enumerate}

 BIC is an alternative metric used for model selection that balances goodness-of-fit with model complexity. BIC serves as a scoring function to evaluate and compare different graph structures. The formulation of BIC is given   {in \cite{Schwarz}} as follows:
\[\text{BIC} = -2 \log L + k \log n \] 
where \( L \) is the likelihood of the data given the model, \( k \) is the number of free parameters in the model, and  \( n \) is the number of data points.

\begin{figure}
    \centering
    \includegraphics[width=1\linewidth]{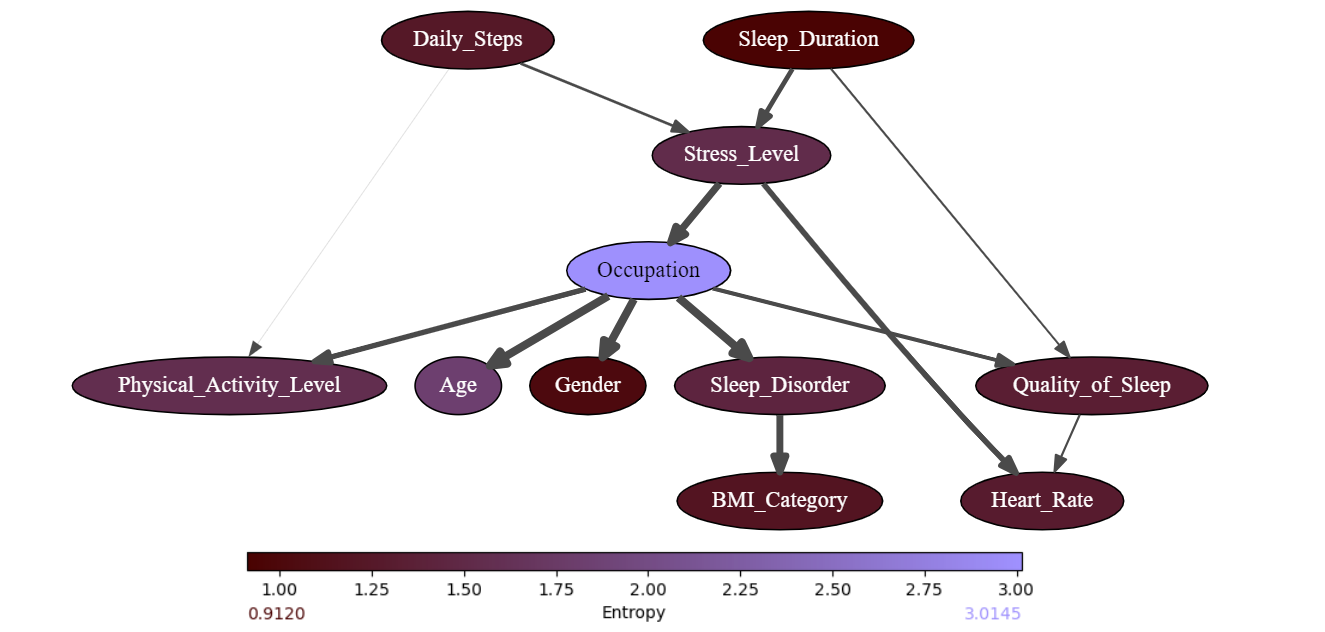}
    \caption{Causal graph structure derived from BIC. Node color indicates entropy values computed per node, while arc thickness represents mutual information calculated for each connection.}
    \label{fig:bic_graph}
\end{figure}

Once we have the structure of the graph (Figure \ref{fig:bic_graph}), we can evaluate it using an independent method such as SEM to ensure that there are statistically significant relationships and to review their logical consistency. The results are displayed in Table \ref{tab:SEM_BIC}.
We observe that most relationships in the causal graph are statistically significant, except between the \textit{Stress level} and \textit{Occupation}. 

\begin{table}[h]
    \centering
    \caption{Relationship between the parent and child nodes in the graph created using LLM. The estimate is the regression coefficient between the variables. The $p$-value  determines whether the relationship is statistically significant; values that show a statistical difference ($p$-value  $\ge$ 0.05) are highlighted in bold}
    \begin{tabular}{l|l|c|c}
        \hline
        \textbf{Child node} & \textbf{Parents node} & \textbf{Estimate} & \textbf{p-value} \\ 
        \hline \hline
        Stress\_Level & Daily\_Steps & 0.5585 & 0 \\
        Stress\_Level & Sleep\_Duration & -1.1091 & 0 \\
        Occupation & Stress\_Level & -0.3907 & \textbf{0.0708} \\
        Quality\_of\_Sleep & Occupation & -0.0281 & 0.0001 \\
        Quality\_of\_Sleep & Sleep\_Duration & 0.7087 & 0 \\
        Sleep\_Disorder & Occupation & -0.1584 & 0 \\
        BMI\_Category & Sleep\_Disorder & -0.4824 & 0 \\
        Heart\_Rate & Stress\_Level & 0.4049 & 0 \\
        Heart\_Rate & Quality\_of\_Sleep & -0.3187 & 0.0004 \\
        Age & Occupation & 0.0588 & 0.0006 \\
        Gender & Occupation & -0.0359 & 0.0001 \\
        Physical\_Activity\_Level & Daily\_Steps & 0.7194 & 0 \\
        Physical\_Activity\_Level & Occupation & 0.0431 & 0.0004 \\
        \hline
    \end{tabular}
    \label{tab:SEM_BIC}
\end{table}

Beyond statistical significance, it is crucial to assess the domain knowledge consistency of the inferred causal relationships. When we examine the directions of causality, we find inconsistencies that suggest potential misinterpretations of causality.
The graph indicates that stress level affects occupation, but logically, it is more reasonable to assume that work-related factors influence stress levels rather than the other way around.
The graph also suggests that variable '\textit{Occupation}' impacts '\textit{Age}' and '\textit{Gender}', whereas  '\textit{Age}' and '\textit{Gender}' should rather influence occupational choices, not the other way around.

These observations highlight an important limitation: while BIC can identify dependencies in data, they do not always correspond to true cause-and-effect relationships. Statistical associations can be influenced by confounding factors, data biases, or incorrect causal assumptions.

\subsection{Causal graph structure based on LLM}

This part of the experiment focuses on generating a causal graph using an LLM as an expert elicitation. This approach leverages the LLM's ability to interpret domain knowledge and reason about causal relationships between variables. 

To implement this approach, we utilized GPT-4o  by OpenAI, and Claude by Anthropic. These models provide advanced reasoning capabilities and extensive knowledge, making them well-suited for analyzing causal relationships. In our methodology, one model generates primary dependencies between variables and finds logical connections between them, while the other verifies the results and points out possible inconsistencies. We effectively create a dual-expert elicitation, where two independent AI systems collaborate to refine the causal graph structure.

\begin{table}[!ht]
    \centering
    \caption{Prompt template for creating the first LLM step to find causal structure}
    \begin{tabular}{p{0.9\textwidth}}  
    \hline 
        \textbf{LLM Prompt Template for Expert Elicitation} \\ \hline \hline 
        
        You are an expert in causal inference and domain knowledge in \textcolor{blue}{[insert domain/topic, e.g., sleep health, education, etc.]}. Your task is to analyze and interpret the causal relationships between variables in the \textcolor{blue}{[insert dataset name]}, leveraging both statistical results and domain expertise. The dataset comprises \textcolor{blue}{[insert number of rows and columns]} and includes the following variables: \textcolor{blue}{[list variables]}.
\\
\\
        Using \textcolor{blue}{[insert causal discovery method]}, causal discovery has been conducted on \textcolor{blue}{[insert details about the dataset]}. The results suggest potential direct causal relationships between certain variables, such as:
\\
        Variable A $->$ Variable B (e.g., [example relationship]).
\\
\\
        Your task is to:

        - Interpret the statistically suggested causal relationships from a domain knowledge perspective.
        
        - Assess the plausibility of these relationships in the context of \textcolor{blue}{[insert domain/topic]}.
        
        - Provide a reasoned explanation for why these relationships may or may not be natural or expected, considering relevant factors such as \textcolor{blue}{[insert relevant factors]}.
\\
\\
        Your response should be detailed, evidence-based, and grounded in expert knowledge of the domain. Consider the interplay between variables and provide a nuanced interpretation of the causal discovery results.
        
        \\ \hline 
    \end{tabular}
    
    \label{tab:LLM_prompt}
\end{table}

To construct a primary causal graph using an LLM, we provide a detailed prompt (Table \ref{tab:LLM_prompt}).
To enhance the expert elicitation process, we created a second prompt (Table \ref{tab:LLM_prompt_2}) that uses the outputs from the first step and performs additional analysis. The second model evaluates the initial causal relationships, identifies potential confounding factors, and assesses the logical consistency of the proposed causal structure.

\begin{table}
    \centering
    \caption{Prompt template for creating expert verification}
    \begin{tabular}{p{0.9\textwidth}}  
    \hline 
        \textbf{LLM Prompt Template for Verification of Finding Relationship} \\ \hline \hline 
        
You are an expert in causal inference and domain knowledge in \textcolor{blue}{[insert domain/topic, e.g., sleep health, education, etc.]}. Your task is to verify and critically evaluate the causal relationships proposed by another expert.
 \\  \\
The first expert proposed the following causal relationships:

{Variable A → Variable B: [Brief description of the relationship].}
 \\  \\
Your task is to:

- Assess the Plausibility. Evaluate whether each proposed relationship is plausible based on your domain knowledge. 

- Identify Confounding Factors. Highlight any potential confounding factors or alternative explanations that could influence the observed relationships.

-  If a relationship seems incorrect or incomplete, suggest corrections or additional relationships that should be considered.
        \\ \hline 
    \end{tabular}
    \label{tab:LLM_prompt_2}
\end{table}

By analyzing the outputs from the second LLM, we identified several confounding variables that were not present in our dataset, but could significantly impact the relationships between the observed variables. These variables include:

\begin{itemize}
    \item Psychological well-being, such as depression, may influence multiple variables, including sleep duration, stress levels, and physical activity.
    \item The ability to adjust work schedules may affect stress levels and sleep patterns.
    \item Socioeconomic status and income may influence stress levels and work conditions.
\end{itemize}

A key challenge in causal graph learning is determining the correct direction of causal relationships. During the LLM-based analysis, we identified several bidirectional dependencies, such as:
\begin{itemize}
    \item \textit{Sleep Duration} $\leftrightarrow$ \textit{Stress Level}
    \item \textit{Heart Rate} $\leftrightarrow$ \textit{Stress Level}
\end{itemize}

\begin{figure}
    \centering
    \includegraphics[width=1\linewidth]{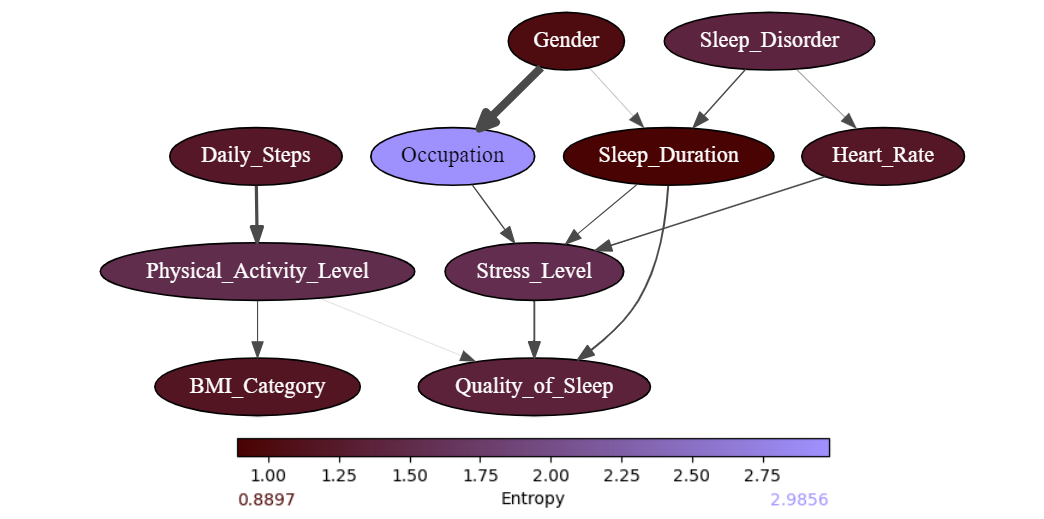}
    \caption{Causal graph structure derived from LLM. Node color indicates entropy values computed per node, while arc thickness represents mutual information calculated for each connection.}
    \label{fig:llm_graph}
\end{figure}

These relationships suggest mutual influence, where each variable may affect the other. However, creating a BN requires to construct a DAG, in which cyclic relationships are not permitted. To resolve these bidirectional connections, we used logical reasoning and domain knowledge to determine the most likely causal direction. 
Overall, 10 out of 12 relationships proposed by the first LLMs were confirmed by the second models, providing a high degree of confidence in the deduced structure.  The final structure is illustrated in Figure \ref{fig:llm_graph}.

Similarly to structure verification based on information criteria, we verified the LLM-base structure using SEM. The results are displayed in the Table \ref{tab:SEM_LLM}

\begin{table}[h]
    \centering
    \caption{Relationship between the parent and child nodes in the graph created using LLM. The estimate is the regression coefficient between the variables. The $p$-value  determines whether the relationship is statistically significant; values that show a statistical difference ($P$-value  $\ge$ 0.05) are highlighted in bold}
    \begin{tabular}{l|l|c|c}
        \hline
        \textbf{Child node} & \textbf{Parents node} & \textbf{Estimate} & \textbf{p-value} \\ 
        \hline \hline
        Physical\_Activity & Daily\_Steps & 0.6582 & 0 \\
        Occupation & Gender & -1.3380 & 0.0001 \\
        Stress\_Level & Sleep\_Duration & -0.7539 & 0 \\
        Stress\_Level & Occupation & -0.0475 & 0.0008 \\
        Stress\_Level & Heart\_Rate & 0.6184 & 0 \\
        Heart\_Rate & Sleep\_Disorder & -0.1029 & 0.0034 \\
        Sleep\_Duration & Sleep\_Disorder & 0.2364 & 0 \\
        Sleep\_Duration & Gender & -0.1613 & 0.0001 \\
        Quality\_of\_Sleep & Stress\_Level & -0.2180 & 0.0001 \\
        Quality\_of\_Sleep & Physical\_Activity & 0.0137 & \textbf{0.5989} \\
        Quality\_of\_Sleep & Sleep\_Duration & 0.5249 & 0 \\
        BMI\_Category & Physical\_Activity & 0.1455 & 0.0001 \\
        \hline
    \end{tabular}
    \label{tab:SEM_LLM}
\end{table}

Based on the analysis of the SEM, it can be concluded that all relationships are statistically significant except for one (\textit{Physical Activity Level $\rightarrow$ Quality of Sleep}), although they were identified as causal relationships based on both LLMs' opinions. The main limitation of this analysis is that the number of records in the dataset is limited, and it is possible that the relationships between these two variables are not well represented.

A common challenge of LLMs is that they are prone to hallucinate \cite{Huang} and generate text that is irrelevant to the topic. However, we can partially mitigate this problem by using different types of LLMs that work together. Each model may have its weaknesses, but because they cross-check the results of the previous model, we are able to minimize and detect inconsistencies that arise due to hallucinations.

\subsection{Comparison of entropy across BN generation methods}

We utilize information-theoretical metrics such as entropy to evaluate the BN structure \cite{Abellan,Ciunkiewicz} generated by different methods: LLM, BIC, and human expert  (Fig.\ref{fig:expert_graph}).  Entropy provides an objective way to compare different BN structures, avoiding potential biases in subjective evaluations. Entropy is a measure of uncertainty in a probability distribution that is computed based on the posterior marginal distribution of each node. The entropy \( H(X) \) of a discrete random node \( x_i \) and corresponding probabilities \( P(x_i) \) is defined as:
\[
H(X) = - \sum_{i} P(x_i) \log P(x_i)
\]
Lower entropy means a more structured model with clearer relationships between variables. Such  model effectively extracts knowledge from the data, identifying patterns and dependencies.
In contrast, higher entropy suggests more randomness and less informative structure.  We calculated the entropy for each node and summarized  all values in Table \ref{tab:entropy_stats}, using the following descriptive statistics:   {
    \begin{itemize}
        \item Mean: the average entropy across all nodes.
        \item Minimum (Min): the lowest entropy value observed among all nodes.
        \item 25th Percentile: indicates that 25\% of the nodes have entropy values below this threshold.
        \item Median: the 50\% value, splitting the dataset into two halves.
        \item 75th Percentile: indicates that 75\% of the nodes have entropy values below this threshold.
        \item Maximum (Max): the highest entropy value observed, representing the most uncertain or unstructured relationships in the BN.
    \end{itemize}
These metrics help compare the structural differences across BN models}. 
\begin{table}[h]
    \centering
    \caption{Summary statistics for Entropy across the BN generated by LLM, BIC, and expert knowledge.}
    \begin{tabular}{lccc}
        \hline
        & \textbf{LLM} & \textbf{BIC} & \textbf{Expert}\\
        \hline
        \textbf{Mean}  & 1.4237 & 1.4770 & 1.4773\\
        \textbf{Min}   & 0.8897 & 0.9119 & 0.9282\\
        \textbf{25\%}  & 1.1654 & 1.1919 & 1.1473\\
        \textbf{50\%}  & 1.2884 & 1.3226 & 1.2075\\
        \textbf{75\%}  & 1.4882 & 1.5410 & 1.5555\\
        \textbf{Max}   & 2.9855 & 3.0144 & 3.2357\\
        \hline
    \end{tabular}
    
    \label{tab:entropy_stats}
\end{table}

The results show that LLM-generated BNs have the lowest mean and minimum entropy. However, the median entropy is slightly lower for expert knowledge, suggesting that in some cases, expert-designed networks may be more structured. 
This implies that the LLM-based BN generation leads to the models with an overall lower uncertainty, making it a strong performer against traditional methods such as BIC and expert-designed networks. 

It is also worth noting that an additional expert review of the logic behind the generated graphs revealed a significant drawback in BNs constructed using the BIC. Specifically, these graphs frequently misidentify causal relationships, leading to incorrect cause-effect links and even the presence of backward connections. Such errors can seriously affect the interpretability and reliability of the model.  It makes it difficult to rely on these structures for decision-making. In contrast, the graphs generated by LLMs show fewer logical inconsistencies, which once again confirms the effectiveness of these approaches in causal modelling as expert elicitation to create the causal structure.

\section{Example of using BN for Decision Support}
\label{sec:DS}

In this section, we demonstrate the application of LLM-generated Bayesian networks (BNs) as expert elicitation tools for decision support using the \textit{Sleep Health and Lifestyle Dataset} in smart health contexts.

\begin{figure*}
    \centering
    \includegraphics[width=1\linewidth]{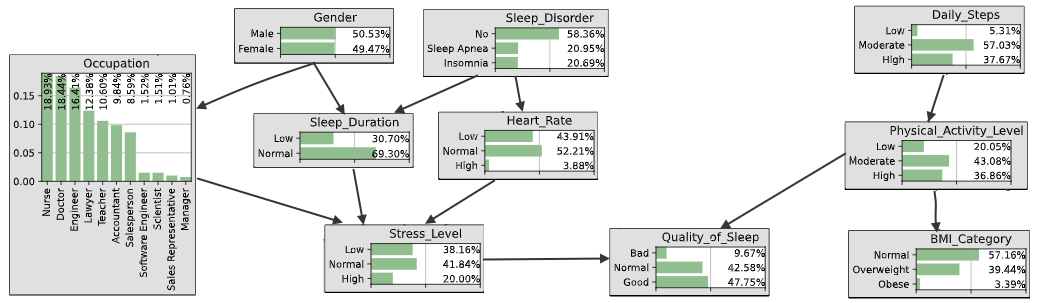}
    \caption{BN generated using LLM expert elicitation  and used as a decision support tool, which represents the relationship between nodes }
    \label{fig:bn_person}
\end{figure*}

\subsection{Creating BN}

To construct a BN, conditional probability tables (CPTs) must be defined for each node. These CPTs quantify data distributions and specify dependencies between variables and their parent nodes, enabling the network to model complex relationships and perform probabilistic inference.

For BN implementation, we employed the PyAgrum library in Python to streamline network creation and inference. The reasoning framework of BNs relies on joint and conditional probabilities, and Bayes formula, which facilitate probabilistic inference.  
 Bayes' formula is given by:
\[
P(A \mid B) = \frac{P(B \mid A) \cdot P(A)}{P(B)}
\]
where \( P(A \mid B) \) is the posterior probability, \( P(B \mid A) \) is the likelihood, \( P(A) \) is the prior, and \( P(B) \) is the marginal probability of the evidence. This probabilistic framework allows BNs to update beliefs dynamically as new evidence becomes available, making them useful tools for reasoning under uncertainty, decision support, and causal analysis.


The sample BN, constructed in this study using the LLMs, consists of a total of 10 nodes, representing variables derived from the dataset   (Fig. \ref{fig:llm_graph}). The distributions between the data in our dataset were used to populate the CPTs.
The BN with the CPTs for assessing stress probabilities, lifestyle and sleep patterns are shown in Figure \ref{fig:bn_person}.

BNs can be used both for inference and diagnostics \cite{Fenton}. Inference in BNs allows us to compute the probability of unknown variables given observed evidence, making predictions and updating beliefs based on new data. This is particularly useful in decision-making under uncertainty, such as assessing the likelihood of a level of stress given the data about the user.
For diagnostics, BNs enable causal analysis by identifying the most probable causes of an observed outcome. 

\subsection{Examples of case-study on the BN}

In this section, we provide two examples of such inferences and diagnostics on the  BN created using the LLM-based expert elicitation.  
  
{The BN nodes represent the variables related to sleep quality, stress levels, and demographic factors. The network captures the relationships between these variables, such as how sleep duration, gender and job type affect stress levels.
We categorized sleep duration into two groups: \textit{"low"} (less than 7 hours) and \textit{"normal"} (7 hours or more). The BN also covers various patterns across 11 occupation types, including healthcare workers: nurses and doctors. This feature enables the model to account for occupation-specific factors, such as work stressors and shift patterns, which can affect sleep and stress levels differently across professions.}

\emph{Example 1. Impact of sleep quality on nurses' stress level}  

Consider a nurse who maintains a normal sleep duration but engages in low physical activity levels. To understand how sleep quality influences stress levels, we examine the probabilities presented in Table \ref{tab:comparison}. If a nurse experiences poor sleep quality, the probability of having a high stress level is 41.56\%, significantly higher compared to those with normal (18.40\%) or good (4.32\%) sleep quality. Conversely, a nurse with good sleep quality has a 92.41\% probability of experiencing low stress, whereas this probability decreases to 26.96\% for those with poor sleep quality. 
These findings suggest strong connections between sleep quality and stress levels. If a nurse does not get appropriate restful sleep, their likelihood of experiencing higher stress increases significantly. 

\begin{table}[h] 
\caption{Probability of nurses' stress level given various levels of sleep quality} 
\label{tab:comparison}
\begin{center}
\begin{tabular}{l|l|l|l}\hline
{Sleep quality}        & {Low, \%}  & {Normal, \%} & {High, \%}\\ \hline \hline
{1. Bad sleep quality}        & {26.96 }  & {31.48 } & {41.56} \\
{2. Normal sleep quality}     & {11.94 }  & {69.67 } & {18.40}\\
{3. Good sleep quality}       & {92.41 }  & {3.27 } & {4.32}\\ \hline
\end{tabular}
\end{center}
\end{table}

\emph{Example 2. Impact of health conditions on stress levels and sleep duration} 

\begin{figure}
    \centering
    \includegraphics[width=0.6\linewidth]{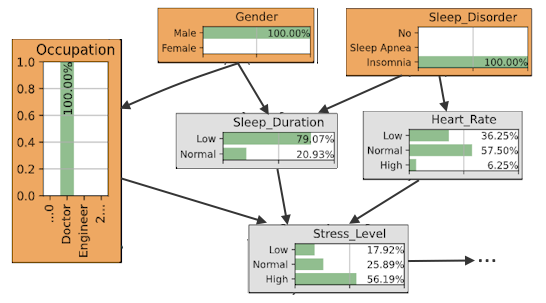}
    \caption{Partial BN showing evidence that the current person is a male doctor with insomnia. The probabilities of stress level and sleep duration are updated accordingly}
    \label{fig:bn_doctor}
\end{figure}

Consider a male doctor who suffers from a sleep disorder such as insomnia and engages in moderate levels of physical activity. To understand how health conditions influence stress levels, we analyze the probabilities from the BN, as partially represented in Figure \ref{fig:bn_doctor}.
Insomnia has a direct impact on sleep duration, significantly increasing the likelihood of short sleep (79.07\%). This, in turn, influences overall stress levels. As shown in our analysis, doctors with insomnia have a 56.19\% probability of experiencing high stress, whereas the chance of maintaining a low stress level drops to just 17.92\%.

\section{Conclusion}

In this study, we explored the use of LLMs as a novel approach for expert elicitation in causal structure learning in healthcare. We compared the LLM-generated BNs to those created using traditional methods like BIC and human expert knowledge. By using entropy and SEM as key metrics, we assessed the quality of the generated BNs, focusing on their ability to capture dependencies while minimizing uncertainty.

 Our findings show that LLM-based BNs tend to have lower overall entropy than those created using BIC and expert input. This suggests that LLMs can serve as a tool for expert elicitation. Also, it is more resilient to wrong conclusions about cause-and-effect connections. It demonstrates fewer logical inconsistencies than those built using the information criterion, often introducing incorrect causal relationships and backward connections. 

The main advantages to using LLM as an expert elicitation include:
\begin{itemize}
    \item \textit{Transparency in elicitation}. Our approach ensures interpretability and explainability in the elicitation process by integrating LLMs. {For example, LLMs can provide justifications and natural language explanations for each proposed causal link, allowing users to trace back and validate the reasoning behind each connection.}
    \item \textit{Accounting for uncertainty}. Experts may have incomplete or uncertain knowledge. We address this by using two different LLMs that verify each other's outputs and causality. This approach helps refine causal relationships.
    \item \textit{Reducing cognitive biases}. Expert judgments can be influenced by biases. Future work should investigate in more detail how the use of LLM can reduce human bias in comparison to a human expert. {In \cite{echterhoff2024}, the authors introduce a framework designed to detect and reduce human-like cognitive biases such as anchoring and sequential effects in LLM outputs. }
\end{itemize}

Thus, the resulting causal model created using LLM has the potential to bring higher confidence of decisions in the systems that utilize BNs. The considered case study demonstrated the feasibility of the LLMs to play the role of experts in the expert elicitation process. Further studies are needed to demonstrate whether applying LLM leads to better explainability of the results of causal modeling and to reduce biases that may be present in the expert human judgments. It may also lead to improving the performance of the causal models in the applications requiring multiple-domain expert elicitation and reducing the cost of the knowledge elicitation process,  compared to the one involving human experts.

\section{Acknowledgment}
This project was partially supported by the Natural Sciences and Engineering Research Council of Canada (NSERC); we also acknowledge scholarships provided by the Department of Biomedical Engineering at the University of Calgary.

\section*{Supplementary Data}
The code created for this study can be accessed via the provided GitHub repository link: \url{https://github.com/ExcellentDarkTea/sleep_health_lifestyle}


\bibliographystyle{IEEEtran}
\bibliography{references} 

\begin{thebibliography}{10}
\providecommand{\url}[1]{#1}
\csname url@samestyle\endcsname
\providecommand{\newblock}{\relax}
\providecommand{\bibinfo}[2]{#2}
\providecommand{\BIBentrySTDinterwordspacing}{\spaceskip=0pt\relax}
\providecommand{\BIBentryALTinterwordstretchfactor}{4}
\providecommand{\BIBentryALTinterwordspacing}{\spaceskip=\fontdimen2\font plus
\BIBentryALTinterwordstretchfactor\fontdimen3\font minus \fontdimen4\font\relax}
\providecommand{\BIBforeignlanguage}[2]{{%
\expandafter\ifx\csname l@#1\endcsname\relax
\typeout{** WARNING: IEEEtran.bst: No hyphenation pattern has been}%
\typeout{** loaded for the language `#1'. Using the pattern for}%
\typeout{** the default language instead.}%
\else
\language=\csname l@#1\endcsname
\fi
#2}}
\providecommand{\BIBdecl}{\relax}
\BIBdecl

\bibitem{Pearl}
J.~Pearl, ``{The Seven Tools of Causal Inference, with Reflections on Machine Learning},'' \emph{{Communications of the ACM}}, vol.~62, no.~3, pp. 54--60, 2019.

\bibitem{Glymour}
C.~Glymour, K.~Zhang, and P.~Spirtes, ``{Review of Causal Discovery Methods Based on Graphical Models},'' \emph{Frontiers in Genetics}, vol.~10, p. 524, 2019.

\bibitem{Darvariu2024}
V.-A. Darvariu, S.~Hailes, and M.~Musolesi, ``{Large Language Models are Effective Priors for Causal Graph Discovery},'' \emph{arXiv preprint arXiv:2405.13551}, 2024.

\bibitem{Takayama}
M.~Takayama, T.~Okuda, T.~Pham, T.~Ikenoue, S.~Fukuma, and S.~Shimizu, ``{Integrating Large Language Models in Causal Discovery: A Statistical Causal Approach},'' \emph{arXiv preprint arXiv:2402.01454}, 2024.

\bibitem{[Bojke-2021]}
L.~Bojke, M.~Soares, K.~Claxton, A.~Colson, A.~Fox, C.~Jackson, D.~Jankovic, A.~Morton, L.~Sharples, and A.~Taylor, ``{Developing a Reference Protocol for Structured Expert Elicitation in Healthcare Decision-making: a Mixed-methods Study},'' \emph{{Health Technology Assessment (Winchester, England)}}, vol.~25, no.~37, p.~1, 2021.

\bibitem{[Cooke-1991]}
R.~Cooke, \emph{{Experts in Uncertainty: Opinion and Subjective Probability in Science}}.\hskip 1em plus 0.5em minus 0.4em\relax Oxford University Press, 1991.

\bibitem{[Mayer_Elicitation_RAND_2021]}
L.~A. Mayer, \emph{{A Structured Elicitation Approach to Identify Technology-Based Challenges: With Application to Inform Force Planning for Technological Surprise}}.\hskip 1em plus 0.5em minus 0.4em\relax RAND, 2021.

\bibitem{Polotskaya}
K.~Polotskaya, C.~S. Mu{\~n}oz-Valencia, A.~Rabasa, J.~A. Quesada-Rico, D.~Orozco-Beltr{\'a}n, and X.~Barber, ``{Bayesian Networks for the Diagnosis and Prognosis of Diseases: A Scoping Review},'' \emph{{Machine Learning and Knowledge Extraction}}, vol.~6, no.~2, pp. 1243--1262, 2024.

\bibitem{McLachlan}
S.~McLachlan, K.~Dube, G.~A. Hitman, N.~E. Fenton, and E.~Kyrimi, ``{Bayesian Networks in Healthcare: Distribution by Medical Condition},'' \emph{{Artificial Intelligence in Medicine}}, vol. 107, p. 101912, 2020.

\bibitem{Shaposhnyk}
O.~Shaposhnyk, S.~Yanushkevich, V.~Babenko, M.~Chernykh, and I.~Nastenko, ``{Inferring Cognitive Load Level from Physiological and Personality Traits},'' in \emph{{2023 International Conference on Information and Digital Technologies (IDT)}}.\hskip 1em plus 0.5em minus 0.4em\relax IEEE, 2023, pp. 233--242.

\bibitem{Fenton}
N.~Fenton and M.~Neil, \emph{{Risk Assessment and Decision Analysis with Bayesian Networks}}.\hskip 1em plus 0.5em minus 0.4em\relax CRC Press, 2018.

\bibitem{Schwarz}
G.~Schwarz, ``{Estimating the Dimension of a Model},'' \emph{The Annals of Statistics}, pp. 461--464, 1978.

\bibitem{Akaike}
H.~Akaike, \emph{{Information Theory and an Extension of the Maximum Likelihood Principle}}, E.~Parzen, K.~Tanabe, and G.~Kitagawa, Eds.\hskip 1em plus 0.5em minus 0.4em\relax Springer New York, 1998.

\bibitem{Spirtes}
P.~Spirtes and C.~Glymour, ``{An Algorithm for Fast Recovery of Sparse Causal Graphs},'' \emph{Social Science Computer Review}, vol.~9, no.~1, pp. 62--72, 1991.

\bibitem{Spirtes-2001}
P.~Spirtes, ``{An Anytime Algorithm for Causal Inference},'' in \emph{International Workshop on Artificial Intelligence and Statistics}.\hskip 1em plus 0.5em minus 0.4em\relax PMLR, 2001, pp. 278--285.

\bibitem{Nogueira}
A.~R. Nogueira, A.~Pugnana, S.~Ruggieri, D.~Pedreschi, and J.~Gama, ``{Methods and Tools for Causal Discovery and Causal Inference},'' \emph{{Wiley Interdisciplinary Reviews: Data Mining and Knowledge Discovery}}, vol.~12, no.~2, p. e1449, 2022.

\bibitem{Versteeg}
P.~Versteeg, J.~Mooij, and C.~Zhang, ``{Local Constraint-based Causal Discovery under Selection Bias},'' in \emph{{Conference on Causal Learning and Reasoning}}.\hskip 1em plus 0.5em minus 0.4em\relax PMLR, 2022, pp. 840--860.

\bibitem{Khatibi}
E.~Khatibi, M.~Abbasian, Z.~Yang, I.~Azimi, and A.~M. Rahmani, ``{ALCM: Autonomous LLM-augmented Causal Discovery Framework},'' \emph{arXiv preprint arXiv:2405.01744}, 2024.

\bibitem{Dataset}
\BIBentryALTinterwordspacing
L.~Tharmalingam, ``{Sleep Health and Lifestyle Dataset}.'' [Online]. Available: \url{https://www.kaggle.com/datasets/uom190346a/sleep-health-and-lifestyle-dataset/data}
\BIBentrySTDinterwordspacing

\bibitem{Shaposhnyk-2023}
O.~Shaposhnyk and S.~Yanushkevich, ``{Integration of Structural Equation Models and Bayesian Networks for Cognitive Load Modeling},'' in \emph{{2023 IEEE Symposium Series on Computational Intelligence (SSCI)}}.\hskip 1em plus 0.5em minus 0.4em\relax IEEE, 2023, pp. 1650--1655.

\bibitem{Verny2017}
L.~Verny, N.~Sella, S.~Affeldt, P.~P. Singh, and H.~Isambert, ``{Learning Causal Networks with Latent Variables from Multivariate Information in Genomic Data},'' \emph{PLoS computational biology}, vol.~13, no.~10, p. e1005662, 2017.

\bibitem{Zamfirescu}
J.~D. Zamfirescu-Pereira, R.~Y. Wong, B.~Hartmann, and Q.~Yang, ``{Why Johnny can’t Prompt: How non-AI Experts Try (and Fail) to Design LLM Prompts},'' in \emph{{Proceedings of the 2023 CHI Conference on Human Factors in Computing Systems}}, 2023, pp. 1--21.

\bibitem{Igolkina}
A.~A. Igolkina and G.~Meshcheryakov, ``{Semopy: A Python Package for Structural Equation Modeling},'' \emph{{Structural Equation Modeling: A Multidisciplinary Journal}}, vol.~27, no.~6, pp. 952--963, 2020.

\bibitem{bollen2014}
K.~A. Bollen, J.~J. Harden, S.~Ray, and J.~Zavisca, ``{BIC and Alternative Bayesian Information Criteria in the Selection of Structural Equation Models},'' \emph{Structural Equation Modeling: A Multidisciplinary Journal}, vol.~21, no.~1, pp. 1--19, 2014.

\bibitem{de2018}
C.~P. de~Campos, M.~Scanagatta, G.~Corani, and M.~Zaffalon, ``{Entropy-based Pruning for Learning Bayesian Networks using BIC},'' \emph{Artificial Intelligence}, vol. 260, pp. 42--50, 2018.

\bibitem{Cabeli}
V.~Cabeli, H.~Li, M.~da~C{\^a}mara Ribeiro-Dantas, F.~Simon, and H.~Isambert, ``{Reliable Causal Discovery based on Mutual Information Supremum Principle for Finite Datasets},'' in \emph{{WHY21 at 35rd Conference on Neural Information Processing Systems NeurIPS}}, 2021.

\bibitem{Huang}
L.~Huang, W.~Yu, W.~Ma, W.~Zhong, Z.~Feng \emph{et~al.}, ``{A Survey on Hallucination in Large Language Models: Principles, Taxonomy, Challenges, and Open Questions},'' \emph{ACM Transactions on Information Systems}, vol.~43, no.~2, pp. 1--55, 2025.

\bibitem{Abellan}
J.~Abell{\'a}n and J.~G. Castellano, ``{Improving the Naive Bayes Classifier via a Quick Variable Selection Method using Maximum of Entropy},'' \emph{Entropy}, vol.~19, no.~6, p. 247, 2017.

\bibitem{Ciunkiewicz}
P.~Ciunkiewicz, M.~Roumeliotis, K.~Stenhouse, P.~McGeachy, S.~Quirk, P.~Grendarova, and S.~Yanushkevich, ``{Assessment of Tissue Toxicity Risk in Breast Radiotherapy using Bayesian Networks},'' \emph{Medical Physics}, vol.~49, no.~6, pp. 3585--3596, 2022.

\bibitem{echterhoff2024}
J.~Echterhoff, Y.~Liu, A.~Alessa, J.~McAuley, and Z.~He, ``{Cognitive Bias in Decision-Making with LLMs},'' \emph{arXiv preprint arXiv:2403.00811}, 2024.

\end{thebibliography}

\end{document}